# Towards an Account of Extraposition in HPSG[*]

**Frank Keller**
Centre for Cognitive Science
University of Edinburgh
2 Buccleuch Place, Edinburgh EH8 9LW, UK
*keller@cogsci.ed.ac.uk*

## Abstract

This paper investigates the syntax of extraposition in the HPSG framework. We present English and German data (partly taken from corpora), and provide an analysis using lexical rules and a nonlocal dependency. The condition for binding this dependency is formulated relative to the antecedent of the extraposed phrase, which entails that no fixed site for extraposition exists. Our analysis accounts for the interaction of extraposition with fronting and coordination, and predicts constraints on multiple extraposition.

## 1 The Data

### 1.1 Extraposition of S and PP

In English, phrases can be extraposed, i.e., dislocated to the right boundary of a sentence. This phenomenon can be observed with adjuncts, such as relative clauses or PPs in (1)–(3), as well as with sentential and prepositional complements as in (4)–(6):[1]

(1) An entirely new band rings today at Great Torrington, [several of whom are members of the congregation]. (UP)

(2) Nobody must live here [who is earning more than twenty pounds a week]. (LL)

(3) A man came into the room [with blond hair]. (CR)

(4) There is very great public concern in Great Britain today (...) [whether the punishments which the courts are empowered to impose are adequate]. (LL)

(5) Extensive and intensive enquiries have been made [into whether this fear of this penalty in fact deters people from murdering]. (LL)

(6) I don't see much argument myself any longer [against differential rents]. (LL)

The antecedent (the category from which the dislocated element is extraposed) is a noun in these cases. Languages in which the right VP boundary is clearly marked (e.g., by the verb if the VP is head-final) can provide evidence for extraposition with verbal antecedents. Cf. the following German data, which include the extraposition of adjuncts in examples (7) and (8), and that of complements in (9) and (10):

(7) In der Nacht hatte es Tote gegeben [in
    in the night had there victims been in
    Moskau (...)]. (STZ)
    Moscow

(8) Er hat den Nerv deutscher Nachkriegs-
    he has the nerve of-German post-war
    geschichte getroffen [mit seiner Roman-
    history hit with his novel
    Triologie (...)]. (STZ)
    trilogy

(9) Aber es wurde öffentlich aufmerksam
    but it was publicly attention
    gemacht [auf eine prekäre Situation]. (STZ)
    called to a delicate situation

(10) Er habe Schipke gesagt, [daß man nicht mit
     he have Schipke said that one not with
     Eiern werfen dürfe, schon gar nicht auf
     eggs throw be-allowed, PART PART not at
     den Bundeskanzler]. (STZ)
     the chancellor

[*]Thanks go to Anette Frank, Tibor Kiss, Jonas Kuhn, Kai Lebeth, and Stefan Müller for comments and suggestions in connection with the research reported here. Part of the work was carried out as part of the Verbmobil Project while the author stayed at the Institute for Logic and Linguistics, IBM Germany, Heidelberg.

[1]Extraposition data was acquired from the following corpora: UPenn Treebank (UP), London-Lund Corpus (LL), Stuttgart Newspaper Corpus (STZ). Other examples were taken from Culicover/Rochemont 1990 (CR), Guéron 1980 (Gue), Haider 1994 (Hai), Nerbonne 1994 (Ner), and Wiltschko 1994 (Wil).

But also in English, we find evidence for extraposition from VP if we assume that adjuncts adjoin to the VP, and hence by default have to follow VP complements:

(11) Florida National said yesterday [that it remains committed to the merger]. (UP)

(12) We're hearing a lot these days [about selling abroad, about the importance of Britain exporting abroad]. (LL)

## 1.2 Multiple Extraposition

It is possible to have more than one extraposed phrase, as shown in (13) and (14):[2]

(13) A man $\_i$ $\_j$ came in [with blond hair]$_i$ [who was smiling]$_j$.

(14) A paper $\_i$ $\_j$ just came out [which talks about extraposition]$_i$ [which you might be interested in]$_j$.

In these examples, both extraposed elements are associated with the same antecedent. We observe that the serialization for multiple extraposed elements matters for PPs, but not for relative clauses:

(15)\* A man $\_i$ $\_j$ came in [who was smiling]$_j$ [with blond hair]$_i$.

(16) A paper $\_i$ $\_j$ just came out [which you might be interested in]$_j$ [which talks about extraposition]$_i$.

We find a different pattern for multiple extraposition involving distinct antecedents:

(17) It$_i$ struck a grammarian $\_j$ last month [who analyzed it]$_j$ [that this clause is grammatical]$_i$. (Hai)

(18)\* It$_i$ struck a grammarian $\_j$ last month [that this clause is grammatical]$_i$ [who analyzed it]$_j$. (Hai)

(19) No one $\_i$ puts things $\_j$ in the sink [that would block it]$_j$ [who wants to go on being a friend of mine]$_i$. (Gue)

(20)\* No one $\_i$ puts things $\_j$ in the sink [who wants to go on being a friend of mine]$_i$ [that would block it]$_j$. (Gue)

It is plausible to assume that multiple extraposition with distinct antecedents is subject to a **nesting requirement**: The first extraposed phrase has to be associated with the last antecedent, the second one with the next-to-last antecedent, etc.

Both types of constraints also apply for German, cf. Wiltschko (1994), who provides extensive evidence for the nesting requirement, including the following data:

(21) weil das Argument $\_i$ einen Mann $\_j$
because the argument a man
aufgeregt hat, [der das Fest besuchte]$_j$ [daß
upset has who the party visited that
Rauchen ungesund ist]$_i$. (Wil)
smoking unhealthy is

(22)\* weil das Argument $\_i$ einen Mann $\_j$ aufgeregt hat, [daß Rauchen ungesund ist]$_i$, [der das Fest besuchte]$_j$ (Wil)

## 1.3 Extraposition and Fronting

The constraint of **frozenness to further extraction**, which states that no dislocation is possible out of an extraposed phrase, is widely accepted in the literature. The contrast between (23) and (24) illustrates this restriction:

(23) Who$_i$ did you see a picture of $\_i$ in the newspaper?

(24)\* Who$_i$ did you see a picture in the newspaper of $\_i$?

Although this constraint seems to be valid for English, it is possible in German to have fronting of material from an extraposed phrase:[3]

(25) Wen$_i$ hast du geglaubt, daß Maria $\_i$
who have you believed that Maria
geküßt hat? (Wil)
kissed has

(26) [Die Maria]$_i$ hat Peter einem Mann $\_j$
the Maria has Peter to-a man
gesagt, [den er kannte]$_j$ [daß er $\_i$ geküßt
said who he knew that he kissed
hat]. (Wil)
has

On the other hand, we can also observe extraposition from fronted phrases, as (27) and (28) show for fronted subjects and objects, respectively.

(27) [Ein Buch $\_j$]$_i$ hat er $\_i$ geschrieben [das
a book has he written which
ihn weltberühmt gemacht hat]$_j$.
him world-famous made has.

(28) [Ein Buch $\_j$]$_i$ war $\_i$ erschienen, [das ihn
a book had appeared which him
weltberühmt gemacht hat]$_j$.
world-famous made has.

We find similar data with extraposition from fronted objects in English:

(29) [Which book $\_j$]$_i$ did she write $\_i$ last year [that takes only two hours to read]$_j$?

(30) [Which woman $\_j$]$_i$ did he meet $\_i$ yesterday [from the south of France]$_j$?

Therefore, we conclude that the phrase structure for extraposition cannot involve a hierarchi-cal constraint which states that extraposed elements are generally higher than fronted ones or vice versa. This is confirmed by the observation that fronted elements can be involved in multiple extraposition as in (26). Our analysis reflects this by avoiding the stipulation of a fixed location for extraposition.

---

[2]We use a trace-like notation to indicate the dependencies with extraposition and fronting phenomena. However, our account of extraposition involves no traces, cf. below.

[3]These examples are less acceptable to speakers of northern variants of German.

### 1.4 Islands and Boundedness

Another common assumption is that extraposition is not subject to the **islands constraints** that hold for extraction to the left. The contrast between (3) and (31) makes clear that subjects are boundaries for fronting, but not for extraposition:

(31)*[With what color hair]$_i$ did a man $\_i$ come into the room? (CR)

Further, the restriction of **upward boundedness** applies to extraposition, i.e., in contrast to fronting, extraposition may not cross the sentence boundary:

(32) Who$_i$ did Mary say [$_S$ that John saw a picture of $\_i$ in the newspaper]? (CR)

(33)*It was believed [$_S$ that John saw a picture $\_i$ in the newspaper by everyone] [of his brother]$_i$. (CR)

We take both constraints as evidence that extraposition is different from fronting and should be handled using a separate nonlocal feature.

## 2 An HPSG Account

### 2.1 Nonlocal Dependencies

We treat extraposition as a nonlocal dependency and introduce a new nonlocal feature EXTRA to establish the connection between an extraposed element and its antecedent.[4] A lexical rule is employed which removes prepositional or verbal complements from the SUBCAT list and introduces them into the EXTRA set:

**Complement Extraposition Lexical Rule (CELR)**

$$\begin{bmatrix} \text{SUBCAT} \boxed{1} \oplus \left\langle \begin{bmatrix} \text{LOC} \boxed{4} | \text{CAT} \begin{bmatrix} \text{HEAD } verb \vee prep \\ \text{SUBCAT} \langle \rangle \end{bmatrix} \end{bmatrix} \right\rangle \oplus \boxed{2} \\ \text{NONLOC} | \text{INHER} | \text{EXTRA} \boxed{3} \end{bmatrix}$$

$$\Longrightarrow \begin{bmatrix} \text{SUBCAT} \boxed{1} \oplus \boxed{2} \\ \text{NONLOC} | \text{INHER} | \text{EXTRA} \boxed{3} \cup \{\boxed{4}\} \end{bmatrix}$$

A similar rule is used to introduce adjuncts into EXTRA:[5]

**Adjunct Extraposition Lexical Rule (AELR)**

$$\begin{bmatrix} \text{LOC} \boxed{2} | \text{CAT} | \text{HEAD } noun \vee verb \\ \text{NONLOC} | \text{INHER} | \text{EXTRA} \boxed{1} \end{bmatrix} \Longrightarrow$$

$$\begin{bmatrix} \text{LOC} | \text{CONT} \boxed{3} \\ \text{NLOC} | \text{INH} | \text{EX} \boxed{1} \cup \left\{ \begin{bmatrix} \text{CAT} \begin{bmatrix} \text{HD } prep \vee rel \begin{bmatrix} \text{MOD} | \text{LOC} \boxed{2} \end{bmatrix} \\ \text{SUBCAT} \langle \rangle \end{bmatrix} \\ \text{CONT} \boxed{3} \end{bmatrix} \right\} \end{bmatrix}$$

Note that the semantic contribution of the adjunct (standardly dealt with by the Semantics Principle) is incorporated into this lexical rule. The sharing $\boxed{3}$ states that the CONT-value of the output is identical with the CONT of the extraposed element, which in turn incorporates the semantics of the input via the sharing $\boxed{2}$.

### 2.2 Periphery Marking

Intuitively, our approach to the phrase structure of extraposition can be formulated as follows: An extraposed constituent has to be bound on top of a phrase that introduces intervening material between the extraposed constituent and its antecedent.[6] Since this constraint on the binding of an extraposed element is relative to its antecedent, we have no fixed site for extraposition, which explains the observed interaction between extraposition and fronting. It also entails a nesting requirement for multiple extraposition, as it triggers distinct binding sites for extraposition from distinct antecedents: The binding site reflects the relative position of the antecedent. Furthermore, we avoid spurious ambiguities which have been problematic for previous accounts.[7]

Our requirement for EXTRA binding can be formulated in HPSG using the notion of **periphery**, which is defined for phrases containing an EXTRA element: A phrase has a **left** periphery iff it contains an EXTRA element which is inherited (a) from its phrasal rightmost daughter or (b) from its lexical head. Otherwise, the phrase has a **right** periphery, and EXTRA elements can be bound on top of it.

---

[4]We have to point out that the use of a nonlocal feature is not crucial to our analysis (as extraposition cannot cross the sentence boundary), but was chosen for technical convenience. Defining EXTRA in this way, we can rely on the Nonlocal Feature Principle for percolation; no additional mechanism is required.

[5]Note that this is a recursive lexical rule, which is rather unusual in standard HPSG. But cf. van Noord/Bouma (1994) who show some other cases where recursive lexical rules are useful and deal with processing issues as well.

[6]Our analysis is inspired by the **Locality Constraint for Identification** (LCI) which Wiltschko (1994) proposes to account for extraposition in a GB framework. The LCI requires that an extraposed element is adjoined at the first maximal projection which dominates its antecedent.

[7]Cf. Keller 1994, where we posited the S node as a fixed site for the binding of extraposed elements. Apart from leading to spurious ambiguities, this assumption is incompatible with the coordination data given in sec. 3.1.

In case (a), no material exists to the right of the extraposed element which could intervene between it and an antecedent. In case (b), the EXTRA element originates directly from a lexical head and would be indistinguishable from a non-extraposed complement or adjunct if bound immediately. Intuitively, in both cases, the EXTRA element has to percolate further up the tree to find a phrase with a right periphery, i.e., one providing intervening material which identifies the EXTRA element as extraposed.

Our periphery definition entails that in a sentence which contains more than one projection with a right periphery, multiple locations for extraposition exist correspondingly. If a sentence contains no projection with a right periphery, no extraposition is possible.

To formalize the notion of periphery, we introduce a new feature PERIPHERY (PER), which is located under LOCAL. Its value is of type *periphery*, defined as follows:

(34) Partition of *periphery*: *extra*, *non-extra*
   Partition of *non-extra*: *left*, *right*

The correct instantiation of PER is guaranteed by the following condition:

(35) **Periphery Marking Condition (PMC)**
   A headed phrase is marked [PER *left*] if it has a daughter D with a non-empty INHER|EXTRA set, and D is
   a. the rightmost daughter and phrasal; or
   b. the head daughter and lexical and marked [PER *left*].

Note that (35b) allows for periphery marking to be specified lexically. We will return to this in sec. 2.6, where we formulate a parochial restriction for German. For English, however, we assume that all lexical entries are marked [PER *left*].

### 2.3 Phrase Structure

To implement the binding of extraposed elements, we introduce an additional immediate dominance schema, which draws on a new subtype of *head-struc* called *head-extra-struc* bearing the feature EXTRA-DTRS (taking a list of *sign*). As the binding of extraposed elements is only possible at the right periphery of a phrase, the head-extra schema specifies its head daughter as [PER *right*] and marks its mother node as [PER *extra*] (the latter is needed for the treatment of adjuncts, cf. sec. 2.5):[8]

---

[8] Here loc($x$) denotes a function which takes as $x$ a list of *sign* and returns a set of *loc* containing the LOC values of the elements of $x$.

**Head-Extra Schema**

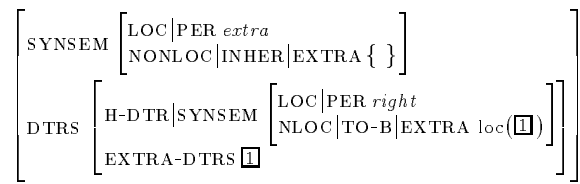

Note that the specification [INHER|EXTRA { }] requires all members of EXTRA to be bound at the same level. This ensures that extraposed elements originating from the same phrase are sisters, and hence can be ordered by LPCs. We use LPCs to account for multiple extraposition from the same antecedent (cf. the data in (13)–(16)):

(36) a. H < E
    b. E [HEAD *prep*] < E [HEAD *verb* ∨ *rel*]

The constraint in (36a) orders the EXTRA-DTRS (E) after the HEAD-DTR (H). With regard to the EXTRA-DTRS, PPs have to precede sentences or relative clauses, as stated in (36b).

### 2.4 Examples

The (simplified) tree structures for (3) and (6) are given in (37) and (38):

(37)
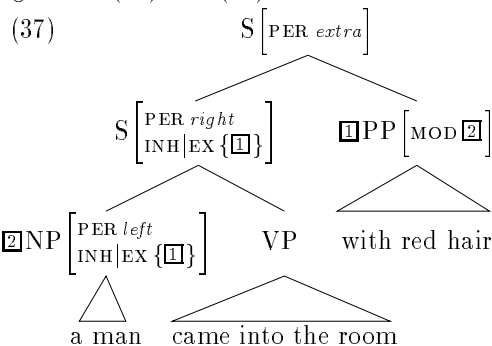

(38)
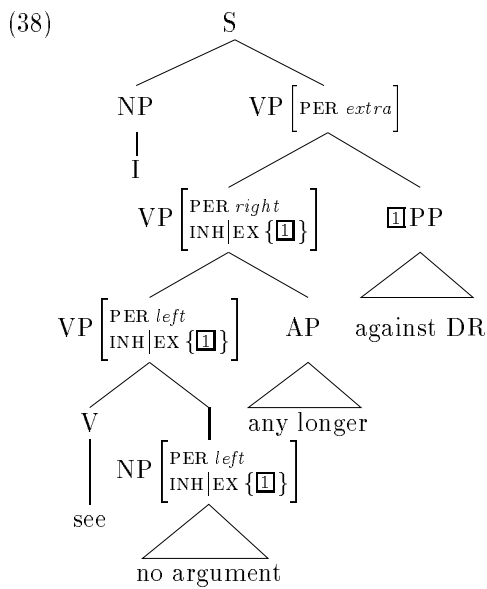

## 2.5 Adjuncts

The phrase structure for extraposition outlined so far has to be constrained further, since it allows adjuncts to adjoin higher than extraposed elements, which is clearly wrong. Cf. the following example with extraposition from NP:

(39)*An entirely new band rings today, [several of whom are members of the congregation] at Great Torrington.

We conclude that the application of the head-adjunct schema has to be disallowed on top of a head-extra structure. This can be achieved straightforwardly by specifying adjuncts as [MOD|LOC|PER *non-extra*].

## 2.6 Extraposition from VP

The AELR has to be restricted language-specifically to account correctly for extraposition from VP:

**English** has a head-initial VP, therefore the right periphery of the VP cannot be formed by the verb, but is provided by VP adjuncts (adverbs and PPs). As a consequence, extraposed VP adjuncts cannot be distinguished from VP adjuncts in base position, which is clearly undesirable. Therefore, we restrict the AELR to nouns on the input side, which disallows adjunct extraposition from VP and hence avoids spurious ambiguities.

In **German**, in contrast, the AELR can apply in full generality. German has a head-final VP, which entails that a verb in final position can form the right periphery of a phrase, making extraposition of VP adjuncts and complements possible. We make use of the lexical constraint in the PMC in (35b) to allow the binding of extraposed elements on top of verbs in final position, which we assume with Pollard (1990) to be marked [INV −]. We can therefore formulate the following lexical requirement:[9]

(40) [INV −] ⇒ [PER *right*]

All other lexical entries are marked [PER *left*], and hence cannot introduce a right periphery.

## 2.7 Fronting

To account for the differences between English and German concerning the fronting from extraposed elements (cf. (24) vs. (25)) we restrict the head-extra schema as follows:

For **English** we assume that both INHER|SLASH and INHER|EXTRA have to be empty for all elements of EXTRA-DTRS. This guarantees that neither fronting nor further extraposition is possible from extraposed phrases.

For **German** we assume that only INHER|EXTRA has to be empty for all elements of EXTRA-DTRS. Therefore, fronting but not extraposition is allowed from extraposed phrases.

# 3 Predictions and Generalizations

## 3.1 Extraposition and Coordination

The head-extra schema together with the PMC has the consequence that elements extraposed from objects are bound at VP level, whereas extraposition from subjects involves binding at S level, as illustrated in (37) and (38). This is confirmed by the following coordination data, which shows that an element which is extraposed from the subject cannot occur at VP level:

(41) [S Nobody must live here and benefit from income support] [who is earning more than twenty pounds a week].

(42)*Nobody must [VP live here] [who is earning more than twenty pounds a week] and [VP benefit from income support].

We find similar data for German, where the subject of a finite clause is related to the S projection via a SLASH dependency, and therefore the head-extra schema applies on top of the head-filler schema:

(43) [S′ Die Behauptung überrraschte mich und
    the claim surprised me and
    erstaunte Maria], [daß Rauchen ung. ist].
    puzzled Maria that smoking unh. is

(44)*Die Behauptung [S überrraschte mich] [daß Rauchen ung. ist] und [S erstaunte Maria].

The Coordination Principle (Pollard/Sag 1994: 202) requires for coordinate structures that the CAT and NONLOC value of each conjunct daughter is identical to that of the mother. If we add the assumption that the mother is always marked as [PER *right*],[10] then the following data with split antecedents can be accounted for:

(45) Ein Mann äußerte die Behauptung und eine
    a man uttered the claim and a
    Frau leugnete die Tatsache daß Rauchen
    woman denied the fact that smoking
    ungesund ist.
    unhealthy is.

Here EXTRA is shared between the conjuncts and bound at S level. Parallel examples exist for English:

---

[9]A similar rule has to be formulated for verbs with separable prefixes since the prefix marks the right periphery.

[10]Note that this is possible as the PMC is valid only for headed structures. We also draw on the fact that PER is a LOCAL feature.

(46) A man came in and a woman went out who knew each other well. (CR)

### 3.2 NP-internal Extraposition

We also find evidence for extraposed phases within NPs, i.e., examples in which adjuncts precede complements:

(47) In [NP an [interview published yesterday]] [with the Los Angeles Daily News], Mr. Simmons said: "Lockheed is actually just a decoy. (...)" (UP)

(48) "The question" at [NP a closed-door meeting [K mart is scheduled to hold today]] [with analysts] "will be: Why aren't we seeing better improvement in sales?" (UP)

These data are not unexpected in our account, since we posit no fixed location for extraposition, and hence allow that an extraposed NP complement is bound inside the NP itself, provided that an adjunct is present to mark the right periphery of the NP. This is the case in (47) and (48).

### 3.3 VP-internal Extraposition

Much in the same vein as with NP-internal extraposition, our account accommodates cases of VP-internal extraposition, which are possible with fronted partial VPs in German:

(49) [VP Einen Hund füttern], [der Hunger hat],
         a   dog  feed   which hunger has
     wird wohl jeder   dürfen. (Ner)
     will PART everyone be-allowed

(50)*Es wird wohl jeder [VP einen Hund füttern], [der Hunger hat], dürfen. (Ner)

The contrast between (49) and (50) shows that extraposition inside a VP is possible only if the VP is fronted. If we assume with Nerbonne (1994) that partial VPs exist in fronted position, but not in the matrix clause, this contrast is readily predicted by our account. Only in fronting examples like (49), the VP does form a separate constituent and hence does exhibit the periphery marking needed for extraposition.

### 3.4 Generalizations

We sum up the generalizations that are captured by our analysis:

(a) Relative clauses, sentences, and PPs can be extraposed, nouns and verbs can function as antecedents. These category restrictions are subject to crosslinguistic variation, as the specific AELR for English shows (cf. sec. 2.6).

(b) Both extraposition from fronted phrases and fronting from extraposed elements are accounted for by our head-extra schema which is constrained by the PMC. In English, fronting from extraposed constituents is disallowed by a language-specific constraint.

(c) The PMC also entails a nesting requirement for extraposed elements with distinct antecedents. Extraposed elements with the same antecedent are bound at the same level and LPCs apply. For English and German, PPs have to precede sentential material. For other languages, different orderings may be stated.

(d) The fact that no island constraints for extraposition exist follows from our use of EXTRA: Island restrictions are formulated for SLASH and hence do not apply to extraposition.

(e) The upward boundedness of extraposition can be captured by stating that a sentence has to be [INHER|EXTRA { }].

(f) Our analysis predicts the asymmetry between extraposition from subjects and objects as found e.g. in coordination data.

(g) NP-internal extraposition and extraposition within fronted VPs are captured without the assumption of any further mechanisms.